\documentstyle[twoside,multicol,psfig]{article}

\catcode`\@=11
\long\def\@makefntext#1{
\protect\noindent \hbox to 3.2pt {\hskip-.9pt
$^{{\footnotesize\@thefnmark}}$\hfil}#1\hfill}		

\def\@makefnmark{\hbox to 0pt{$^{\@thefnmark}$\hss}}	
	
\def\ps@myheadings{%
    \let\@oddfoot\@empty\let\@evenfoot\@empty
    \def\@evenhead{\footnotesize\it\leftmark\hfil}
    \def\@oddhead{\hfil{\footnotesize\it\rightmark}}
    \let\@mkboth\@gobbletwo
    \let\sectionmark\@gobble
    \let\subsectionmark\@gobble
    }

\oddsidemargin=\evensidemargin
\newcounter{sectionc}\newcounter{subsectionc}\newcounter{subsubsectionc}
\renewcommand{\section}[1] {\vspace{14pt}\addtocounter{sectionc}{1}
\setcounter{subsectionc}{0}\setcounter{subsubsectionc}{0}\noindent
	{\bf\thesectionc. #1}\par\vspace{8pt}}
\renewcommand{\subsection}[1] {\vspace{14pt}\addtocounter{subsectionc}{1}
   \setcounter{subsubsectionc}{0}\noindent
   {\bf\thesectionc.\thesubsectionc. {\kern1pt \bfit #1}}\par\vspace{8pt}}
\renewcommand{\subsubsection}[1] {\vspace{14pt}
    \addtocounter{subsubsectionc}{1}
	\noindent{\thesectionc.\thesubsectionc.\thesubsubsectionc.
	{\kern1pt \it #1}}\par\vspace{8pt}}
\newcommand{\nonumsection}[1] {\vspace{14pt}\noindent{\bf #1}
	\par\vspace{8pt}}

\newcounter{appendixc}
\newcounter{subappendixc}[appendixc]
\newcounter{subsubappendixc}[subappendixc]
\renewcommand{\thesubappendixc}{\Alph{appendixc}.\arabic{subappendixc}}
\renewcommand{\thesubsubappendixc}
	{\Alph{appendixc}.\arabic{subappendixc}.\arabic{subsubappendixc}}

\renewcommand{\appendix}[1] {\vspace{14pt}
        \refstepcounter{appendixc}
        \setcounter{figure}{0}
        \setcounter{table}{0}
        \setcounter{lemma}{0}
        \setcounter{theorem}{0}
        \setcounter{corollary}{0}
        \setcounter{definition}{0}
        \setcounter{equation}{0}
        \renewcommand{\thefigure}{\Alph{appendixc}.\arabic{figure}}
        \renewcommand{\thetable}{\Alph{appendixc}.\arabic{table}}
        \renewcommand{\theappendixc}{\Alph{appendixc}}
        \renewcommand{\thelemma}{\Alph{appendixc}.\arabic{lemma}}
        \renewcommand{\thetheorem}{\Alph{appendixc}.\arabic{theorem}}
        \renewcommand{\thedefinition}{\Alph{appendixc}.\arabic{definition}}
        \renewcommand{\thecorollary}{\Alph{appendixc}.\arabic{corollary}}
        \renewcommand{\theequation}{\Alph{appendixc}.\arabic{equation}}
        \noindent{\bf Appendix \theappendixc #1}\par\vspace{5pt}}
\newcommand{\subappendix}[1] {\vspace{14pt}
        \refstepcounter{subappendixc}
        \noindent{\bf Appendix \thesubappendixc. {\kern1pt \bfit #1}}
	\par\vspace{8pt}}
\newcommand{\subsubappendix}[1] {\vspace{14pt}
        \refstepcounter{subsubappendixc}
        \noindent{\rm Appendix \thesubsubappendixc. {\kern1pt \it #1}}
	\par\vspace{8pt}}

\newcommand{\textlineskip}{\baselineskip=13pt}
\newcommand{\smalllineskip}{\baselineskip=10pt}

\newcommand{\copyrightheading}[1]
	{\vspace*{-2.5cm}\smalllineskip{\flushleft
	{\footnotesize International Journal of Neural Systems, #1}\\
	{\footnotesize \copyright\, World Scientific Publishing
	 Company}\\
	 }}

\newcommand{\publisher}[2]{{\begin{center}\tenrm\baselineskip=12pt
	Received #1\\
	Revised #2
	\end{center}
	}}

\def\abstracts#1#2{{
	\centering{\begin{minipage}{5.8in}\small\baselineskip=11pt
	\parindent=0pc #1\par
	\parindent=2pc #2
	\end{minipage}}\par}}

\renewenvironment{thebibliography}[1]		
	{\small\baselineskip=11pt
	 \frenchspacing
	 \begin{list}{\arabic{enumi}.}
        {\usecounter{enumi}\setlength{\parsep}{0pt}
	 \setlength{\leftmargin 12.7pt}{\rightmargin 0pt}
         \setlength{\itemsep}{0pt} \settowidth
	{\labelwidth}{#1.}\sloppy}}{\end{list}}

\newcounter{itemlistc}
\newcounter{romanlistc}
\newcounter{alphlistc}
\newcounter{arabiclistc}

\newcommand{\fcaption}[1]{
        \refstepcounter{figure}
        \setbox\@tempboxa = \hbox{\small Fig.~\thefigure. #1}
        \ifdim \wd\@tempboxa > 5in
           {\begin{center}
        \parbox{5in}{\small\baselineskip=11pt Fig.~\thefigure. #1}
            \end{center}}
        \else
             {\begin{center}
             {\small Fig.~\thefigure. #1}
              \end{center}}
        \fi}

\newcommand{\tcap}[1]{
        \refstepcounter{table}
        \setbox\@tempboxa = \hbox{\small Table~\thetable. #1}
        \ifdim \wd\@tempboxa > 3.15in
           {\begin{center}
        \parbox{3.15in}{\small\smalllineskip
	    Table~\thetable. #1}
            \end{center}}
        \else
             {\begin{center}
             {\eightpoint Table~\thetable. #1}
              \end{center}}
        \fi}

\def\@citex[#1]#2{\if@filesw\immediate\write\@auxout
	{\string\citation{#2}}\fi
\def\@citea{}\@cite{\@for\@citeb:=#2\do
	{\@citea\def\@citea{,}\@ifundefined
	{b@\@citeb}{{\bf ?}\@warning
	{Citation `\@citeb' on page \thepage \space undefined}}
	{\csname b@\@citeb\endcsname}}}{#1}}

\newif\if@cghi
\def\cite{\@cghitrue\@ifnextchar [{\@tempswatrue
	\@citex}{\@tempswafalse\@citex[]}}
\def\citelow{\@cghifalse\@ifnextchar [{\@tempswatrue
	\@citex}{\@tempswafalse\@citex[]}}
\def\@cite#1#2{{$\null^{#1}$\if@tempswa\typeout
	{IJCGA warning: optional citation argument
	ignored: `#2'} \fi}}

\def\pmb#1{\setbox0=\hbox{#1}
	\kern-.025em\copy0\kern-\wd0
	\kern.05em\copy0\kern-\wd0
	\kern-.025em\raise.0433em\box0}

\def\fnt#1#2{\footnotetext{\kern-.3em
	{$^{\mbox{\scriptsize #1}}$}{#2}}}

\font\tenrm=cmr10
\font\tenit=cmti10

\font\bfit=cmbxti10 at 10pt

\font\eightit=cmti8

\def\itlatex{\tenit L\kern-.30em\raise.4ex\hbox{\eightit A}\kern-.14em
T\kern-.1667em\lower.7ex\hbox{E}\kern-.125em X}

\def\bsc{{\sc a\kern-7pt\sc a}}
\def\bflatex{\bf L\kern-.30em\raise.3ex\hbox{\bsc}\kern-.18em
T\kern-.1667em\lower.7ex\hbox{E}\kern-.125em X}

\def\qed{\hbox{${\vcenter{\vbox{			
   \hrule height 0.4pt\hbox{\vrule width 0.4pt height 6pt
   \kern5pt\vrule width 0.4pt}\hrule height 0.4pt}}}$}}

\begin{document}
\setlength{\textheight}{8.78truein}     

\thispagestyle{empty}

\markboth{Multiset CCA for SSVEP Recognition}
{Multiset CCA for SSVEP Recognition}

\textlineskip
\setcounter{page}{1}

\copyrightheading{Vol.~0, No.~0 (XX, 2013) 00--00}

\vspace*{1.05truein}


\centerline{\large\bf FREQUENCY RECOGNITION IN SSVEP-BASED BCI USING}
\vspace*{0.04truein}
\centerline{\large\bf MULTISET CANONICAL CORRELATION ANALYSIS}
\vspace*{0.45truein}
\centerline{YU ZHANG}
\vspace*{0.0215truein}
\centerline{\it Key Laboratory for Advanced Control and Optimization for Chemical Processes}
\vspace*{0.0215truein}
\centerline{\it East China University of Science and Technology, Shanghai, China}
\baselineskip=11pt
\centerline{\it E-mail: yuzhang@ecust.edu.cn}
\vspace*{14pt}
\centerline{GUOXU ZHOU}
\vspace*{0.0215truein}
\centerline{\it Laboratory for Advanced Brain Signal Processing}
\vspace*{0.0215truein}
\centerline{\it RIKEN Brain Science Institute, Wako-shi, Japan}
\baselineskip=11pt
\centerline{\it E-mail: zhouguoxu@brain.riken.jp}
\vspace*{14pt}
\centerline{JING JIN}
\vspace*{0.0215truein}
\centerline{\it Key Laboratory for Advanced Control and Optimization for Chemical Processes}
\vspace*{0.0215truein}
\centerline{\it East China University of Science and Technology, Shanghai, China}
\baselineskip=11pt
\centerline{\it E-mail: jinjingat@gmail.com}
\vspace*{14pt}
\centerline{XINGYU WANG}
\vspace*{0.0215truein}
\centerline{\it Key Laboratory for Advanced Control and Optimization for Chemical Processes}
\vspace*{0.0215truein}
\centerline{\it East China University of Science and Technology, Shanghai, China}
\baselineskip=11pt
\centerline{\it E-mail: xywang@ecust.edu.cn}
\vspace*{14pt}
\centerline{ANDRZEJ CICHOCKI}
\vspace*{0.0215truein}
\centerline{\it Laboratory for Advanced Brain Signal Processing}
\vspace*{0.0215truein}
\centerline{\it RIKEN Brain Science Institute, Wako-shi, Japan}
\vspace*{0.0215truein}
\centerline{\it and}
\vspace*{0.0215truein}
\centerline{\it Systems Research Institute}
\vspace*{0.0215truein}
\centerline{\it Polish Academy of Science, Warsaw, Poland}
\baselineskip=11pt
\centerline{\it E-mail: a.cichocki@riken.jp}

\vspace*{0.3truein}
\publisher{~(to be inserted}{~by Publisher)}

\vspace*{0.29truein}
\abstracts{Canonical correlation analysis (CCA) has been one of the most popular methods for frequency recognition in steady-state visual evoked potential (SSVEP)-based brain-computer interfaces (BCIs). Despite its efficiency, a potential problem is that using pre-constructed sine-cosine waves as the required reference signals in the CCA method often does not result in the optimal recognition accuracy due to their lack of features from the real EEG data. To address this problem, this study proposes a novel method based on multiset canonical correlation analysis (MsetCCA) to optimize the reference signals used in the CCA method for SSVEP frequency recognition. The MsetCCA method learns multiple linear transforms that implement joint spatial filtering to maximize the overall correlation among canonical variates, and hence extracts SSVEP common features from multiple sets of EEG data recorded at the same stimulus frequency. The optimized reference signals are formed by combination of the common features and completely based on training data. Experimental study with EEG data from ten healthy subjects demonstrates that the MsetCCA method improves the recognition accuracy of SSVEP frequency in comparison with the CCA method and other two competing methods (multiway CCA (MwayCCA) and phase constrained CCA (PCCA)), especially for a small number of channels and a short time window length. The superiority indicates that the proposed MsetCCA method is a new promising candidate for frequency recognition in SSVEP-based BCIs.}{}
\vspace*{14pt}
\abstracts{\emph{Keywords}: Brain-computer interface (BCI); Electroencephalogram (EEG); Multiset canonical correlation analysis (MsetCCA); Steady-state visual evoked potential (SSVEP).}{}

\vspace*{10pt}\textlineskip

\begin{multicols}{2}
\section{Introduction}
\noindent
Brain-computer interfaces (BCIs) are communication systems that translate brain electrical activities typically measured by electroencephalogram (EEG) into computer commands, and hence assist to reconstruct communicative and environmental control abilities for severely disabled people \cite{BCI1}$^-$\cite{BCIWang}. Event-related potential (ERP), steady-state visual evoked potential (SSVEP) and event-related (de)synchronization (ERD/ERS) are usually adopted for development of BCIs \cite{ERP-BCI}$^-$\cite{MI-Lee}. In recent years, SSVEP-based BCI has been increasingly studied since it requires less training to the user and provides relatively higher information transfer rate (ITR) \cite{PcodeSSVEP,SSVEP-Allison,SSVEP-Pan}.

When subject focuses attention on the repetitive flicker of a visual stimulus, SSVEP is elicited at the same frequency as the flicker frequency and also its harmonics over occipital scalp areas \cite{SSVEP-Muller,SSVEP-Cheng}. Accordingly, SSVEP-based BCI is designed to detect the desired commands  through recognizing the SSVEP frequency in EEG. Although original SSVEP responses present relatively stable spectrums over time, they are likely to be contaminated by ongoing EEG activities and other background noises \cite{MultiChannel-Friman,SSVEP-CCA}. Therefore, development of an effective algorithm to recognize the SSVEP frequency with a high accuracy and a short time window length (TW) is considerably important for development of an SSVEP-based BCI with high performance.

So far, various approaches have been proposed to recognize SSVEP frequency for BCI applications \cite{SSVEP-PSDA}$^-$\cite{SSVEP-SC}. Among them, a canonical correlation analysis (CCA)-based recognition method \cite{SSVEP-CCA}, first introduced by Lin et al., has aroused more interests of researchers. The CCA method implements correlation maximization between the multichannel EEG signals and the pre-constructed reference signals with sine-cosine waves at each of the used stimulus frequencies. The stimulus frequency corresponding to the maximal correlation coefficient is then recognized as the SSVEP frequency. The CCA method has shown significantly better recognition performance than that of the traditional power spectral density analysis (PSDA) using a single or bipolar channel \cite{SSVEP-CCA,SSVEP-Bin}. Although the CCA method has been validated by many studies on SSVEP-based BCIs \cite{SSVEP-Bin,SSVEP-Zhang}, a potential problem of this approach is that all parameters for recognition are estimated from test data since the reference signals of sine-cosine waves do not include features from training data. Hence, the CCA method often does not result in the optimal recognition accuracy of SSVEP frequency due to possible overfitting, especially using a short TW \cite{MCCA,SSVEP-PCCA}. Recently, a multiway extension of CCA (MwayCCA) \cite{MCCA} has been proposed by Zhang et al. to optimize the reference signals used in the CCA method from multiple dimensions of EEG tensor data for SSVEP frequency recognition. The MwayCCA method has shown improved recognition performance of SSVEP frequency compared to the CCA method. On the other hand, a phase constrained CCA method (PCCA) \cite{SSVEP-PCCA} has also been proposed for SSVEP frequency recognition. The PCCA method achieved significant accuracy improvement in comparison with the CCA method, by embedding the phase information estimated from training data into the reference signals. However, the procedures of reference signal optimization in both the MwayCCA and the PCCA methods are not completely based on training data but still need to resort to the pre-constructed sine-cosine waves.

In the present study, we introduce a multiset canonical correlation analysis (MsetCCA)-based method to reference signal optimization for SSVEP frequency recognition. MsetCCA was developed as an extension of CCA to find multiple linear transforms that maximize the overall correlation among canonical variates from multiple sets of random variables \cite{MsetCCA}. Recently, MsetCCA has been successfully applied to the joint blind source separation of multi-subject fMRI data \cite{JBSSMsetCCA,GroupMsetCCA}, the functional connectivity analysis of fMRI data \cite{RMsetCCA}, and also the fusion of concurrent single trial ERP and fMRI data \cite{MsetCCA-Correa}. In this study, the proposed MsetCCA method implements reference signal optimization for SSVEP frequency recognition through extracting SSVEP common features from the joint spatial filtering of multiple sets of EEG data recorded at the same stimulus frequency. Different from the MwayCCA and the PCCA methods, the procedure of reference signal optimization in the MsetCCA method is completely based on training data. EEG data recorded from ten healthy subjects are used to validate the MsetCCA method in comparison with the CCA, the MwayCCA and the PCCA methods. Experimental results indicate that the proposed MsetCCA method outperforms the three competing methods for SSVEP recognition, especially for a small number of channels and a short time window length.

\section{Materials and Methods}
\vspace*{-1pt}
\subsection{Experiments and EEG recordings}
\begin{figure*}[!t]
\vspace*{13pt}
\centerline{\psfig{file=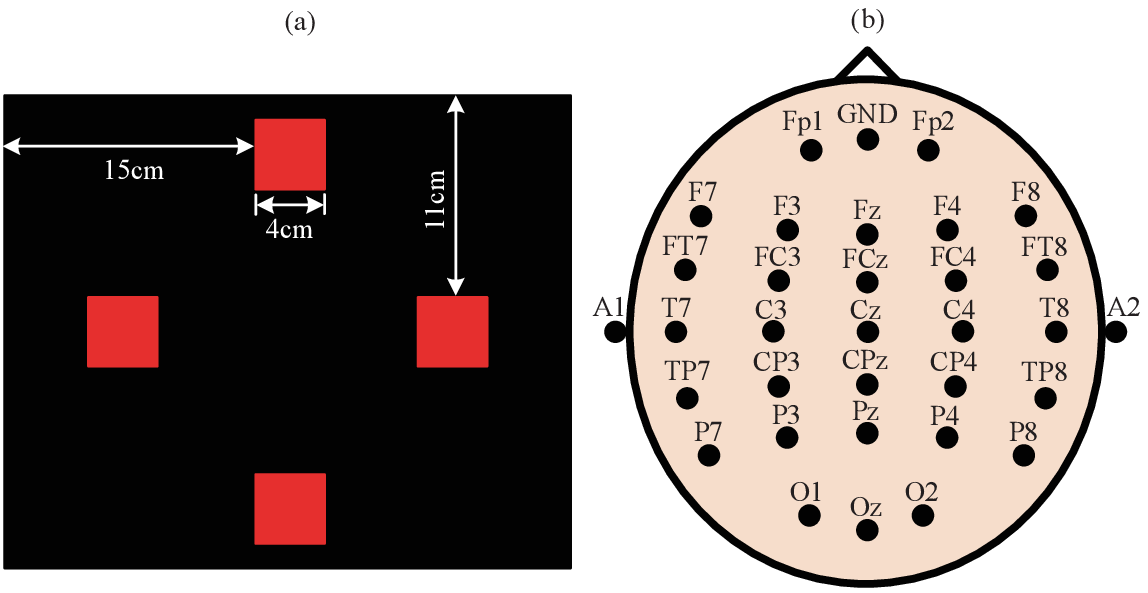,width=0.7\textwidth}}
\fcaption{Experimental layout (a) and channel configuration (b) for EEG recordings.} \label{F-Layout}
\end{figure*}

\vspace*{-4pt}
\noindent
The experiments were performed by ten healthy subjects (S1-S10, aged from 21 to 27, all males) who received their remuneration. All of them had normal or corrected to normal vision. In the experiments, the subjects were seated in a comfortable chair 60 cm from a standard 17 inch CRT monitor (85 Hz refresh rate, 1024 $\times$ 768 screen resolution) in a shielded room. Four red squares were presented on the screen as stimuli (see Fig.~\ref{F-Layout} (a)) and flickered at different four frequencies 6 Hz, 8 Hz, 9 Hz and 10 Hz, respectively. There were 20 experimental runs completed by each subject. In each run, the subject was asked to focus attention on each of the four stimuli once for 4 s, respectively, preceded by each target cue duration of 2 s. A total of 80 trials (4 trials in each run) were therefore performed by each subject.

EEG signals were recorded by using the Nuamps amplifier (NuAmp, Neuroscan, Inc.) at 250 Hz sampling rate with high-pass and low-pass filters of 0.1 and 70 Hz from 30 channels arranged according to standard positions of the 10-20 international system (see Fig.~\ref{F-Layout} (b)). The average of two mastoid electrodes (A1, A2) was used as reference and the electrode on the forehead (GND) as ground. With a sixth-order Butterworth filter, a band-pass filtering from 4 to 45 Hz was implemented on the recorded EEG signals before further analysis.

\vspace*{13pt}

\vspace*{-1pt}
\subsection{CCA for SSVEP Recognition}
\vspace*{-4pt}
As a multivariate statistical method, canonical correlation analysis (CCA) \cite{CCA} explores the underlying correlation between two sets of data. Given two sets of random variables ${\bf X} \in {\bf R}^{I_1 \times J}$ and ${\bf Y} \in {\bf R}^{I_2 \times J}$, which are normalized to have zero mean and unit variance, CCA is to seek a pair of linear transforms ${\bf w}_x \in {\bf R}^{I_1}$ and ${\bf w}_y \in {\bf R}^{I_2}$ such that the correlation between linear combinations ${\bf \tilde x} = {\bf w}_x^T{\bf X}$ and ${\bf \tilde y} = {\bf w}_y^T{\bf Y}$ is maximized as
\begin{eqnarray} \label{E-CCA}
\mathop{\max}\limits_{{\bf w}_x,{\bf w}_y} \rho & = & \frac{E\left[{{\bf \tilde x}{\bf \tilde y}^T}\right]}{\sqrt{E\left[{{\bf \tilde x}{\bf \tilde x}^T}\right]E\left[{{\bf \tilde y}{\bf \tilde y}^T}\right]}} \nonumber  \\
& = & \frac{{\bf w}_x^T{\bf X}{\bf Y}^T{\bf w}_y} {\sqrt{{\bf w}_x^T{\bf X}{\bf X}^T{\bf w}_x{\bf w}_y^T{\bf Y}{\bf Y}^T{\bf w}_y}}.
\end{eqnarray}
The maximum of correlation coefficient $\rho$ with respect to ${\bf w}_x$ and ${\bf w}_y$ is the maximum canonical correlation.

A CCA-based frequency recognition method \cite{SSVEP-CCA} was first introduced by Lin et al. to SSVEP-based BCI. The CCA method provided better recognition performance than that of the PSDA since it delivered an optimization for the combination of multiple channels to improve the signal-to-noise ratio (SNR). Assume our aim is to recognize the target frequency (i.e., SSVEP frequency) from $M$ stimulus frequencies in an SSVEP-based BCI. ${\bf \hat X} \in {\bf R}^{C \times P}$ ($C$ channels $\times$ $P$ time points) is a test data set consisted of EEG signals from $C$ channels with $P$ time points in each channel. ${\bf Y}_m \in {\bf R}^{2H \times P}$ is a pre-constructed reference signal set at the $m$-th stimulus frequency $f_m$ ($m=1,2,\ldots,M$) and is formed by a series of sine-cosine waves as
\begin{equation} \label{E-SinCos}
{\bf Y}_m  = \left( {\begin{array}{*{20}c}
                {\sin \left( {2\pi f_m t} \right)}  \\
                {\cos \left( {2\pi f_m t} \right)}  \\
                \vdots   \\
                {\sin \left( {2\pi Hf_m t} \right)}  \\
                {\cos \left( {2\pi Hf_m t} \right)}  \\
                \end{array}} \right), \quad t = \frac{1}{F},\frac{2}{F},\ldots,\frac{P}{F},
\end{equation}
where $H$ is the number of harmonics and $F$ denotes the sampling rate. Solving the maximal correlation coefficient $\rho_m$ between ${\bf \hat X}$ and ${\bf Y}_m$ $(m=1,2,\ldots,M)$ by (\ref{E-CCA}), the SSVEP frequency is then recognized by
\begin{equation} \label{E-Ft}
\hat f = \mathop{\arg\max}\limits_{f_m}\rho_m, \quad m=1,2,\ldots,M.
\end{equation}

\vspace*{-1pt}
\subsection{MsetCCA for SSVEP Recognition}
\vspace*{-4pt}
Good performance of the CCA-based recognition method has been confirmed by many studies on SSVEP-based BCIs \cite{SSVEP-Bin,SSVEP-Zhang}. However, the pre-constructed reference signals of sine-cosine waves often do not result in the optimal recognition accuracy of SSVEP frequency due to their lack of important information contained in the real EEG data. We consider that some common features should be shared by a set of trials recorded at a certain stimulus frequency on a same subject. Such common features contained in the real EEG data could be more natural reference signals in contrast to sine-cosine waves for SSVEP frequency recognition of a test set. Motivated by this idea, we propose a method based on multiset canonical correlation analysis (MsetCCA) to extract the common features for reference signal optimization as a sophisticated calibration procedure followed by the SSVEP frequency recognition using CCA, and hence to improve the recognition accuracy further. The optimal reference signals are first learned by MsetCCA from the joint spatial filtering of multiple sets of EEG training data for each of the stimulus frequencies, and are subsequently used instead of sine-cosine waves in the CCA method for SSVEP frequency recognition.

MsetCCA is a generalization of CCA to more than two sets of data, which maximizes the overall correlation among canonical variates from multiple sets of random variables through optimizing the objective function of correlation matrix of the canonical variates \cite{JBSSMsetCCA,GroupMsetCCA,MsetCCA-Correa}. The five most discussed objective functions \cite{MsetCCA,MsetCCA-Nielsen} for MsetCCA are: (1) SUMCOR, maximize the sum of all entries in the correlation matrix; (2) SSQCOR, maximize the sum of squares of all entries in the correlation matrix; (3) MAXVAR, maximize the largest eigenvalue of the correlation matrix; (4) MINVAR, minimize the smallest eigenvalue of the correlation matrix; (5) GENVAR, minimize the determinant of the correlation matrix. All of the five objective functions yield similar results on a group dataset \cite{JBSSMsetCCA}. This study adopts the MAXVAR approach to solve MsetCCA, since it provides a natural extension of CCA to multiple sets \cite{RMsetCCA,MaxvarCCA}.

\begin{figure*}[!t]
\vspace*{13pt}
\centerline{\psfig{file=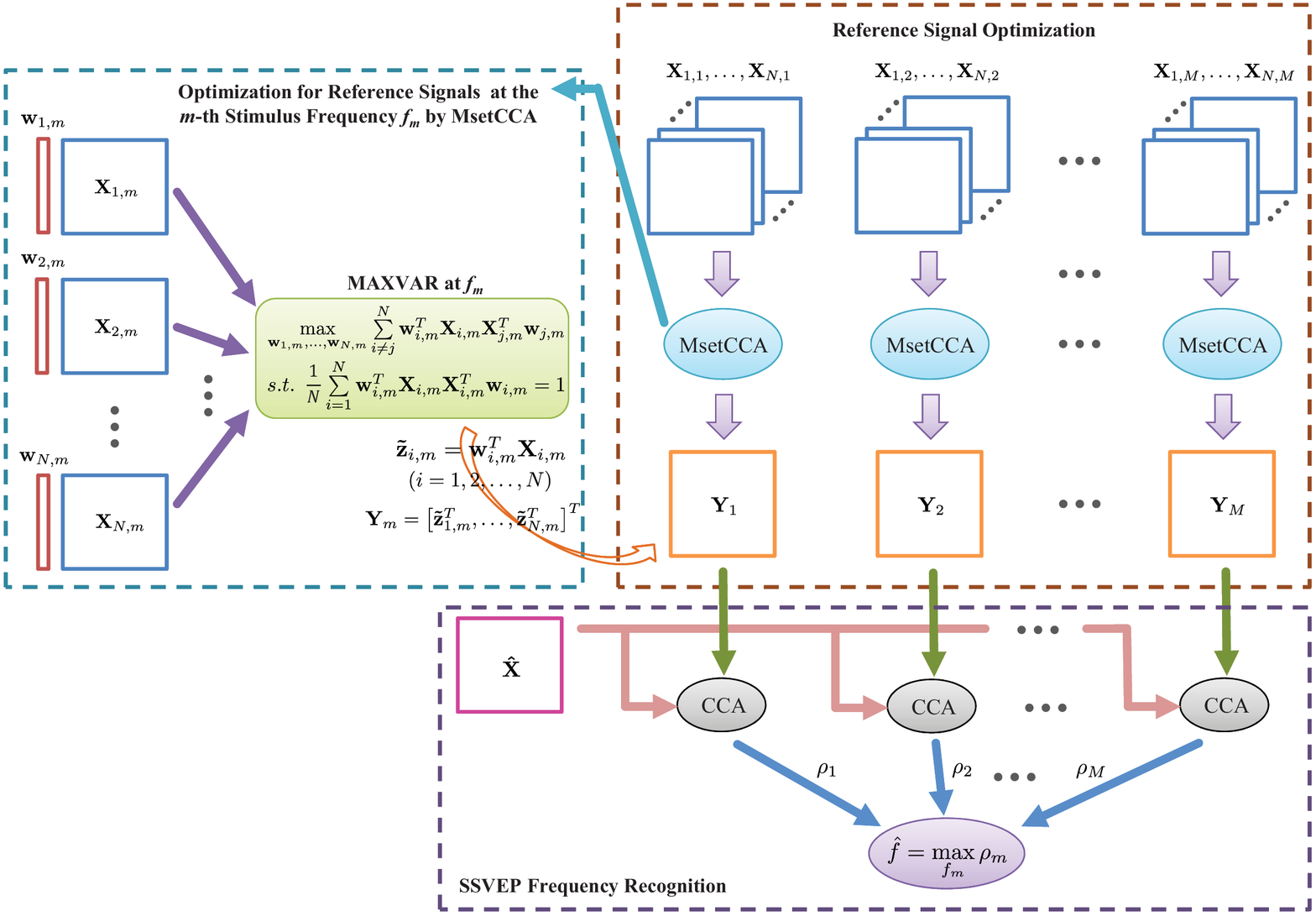,angle=-90,width=0.85\textwidth}}
\vspace*{13pt}
\fcaption{Illustration of the MsetCCA-based method for SSVEP frequency recognition. ${\bf X}_{1,m},\ldots,{\bf X}_{N,m} \in {\bf R}^{C \times P}$ denote the EEG training data sets recorded from $N$ experimental trials at the $m$-th stimulus frequency $f_m$ ($m=1,2,\ldots,M$). Multiple linear transforms (i.e., spatial filters) ${\bf w}_{1,m},\ldots,{\bf w}_{N,m}$ are found by MsetCCA with the objective function MAXVAR on the training data. Canonical variates are then computed through ${\bf \tilde z}_{i,m}={\bf w}_{i,m}^T{\bf X}_{i,m}$ $(i=1,2,\ldots,N)$ and combined as formula (\ref{E-ORS}) to form the optimized reference signal set ${\bf Y}_m \in {\bf R}^{N \times P}$ at $f_m$. Regarding a new test data set ${\bf \hat X} \in {\bf R}^{C \times P}$, the maximal correlation coefficient between ${\bf \hat X}$ and each of ${\bf Y}_m$ ($m=1,2,\ldots,M$) is computed by CCA in formula (\ref{E-CCA}). The SSVEP frequency in $\mathbf{\hat X}$ is then recognized by formula (\ref{E-Ft}).} \label{F-MsetCCAforSSVEP}
\end{figure*}
\vspace*{13pt}

Assume we have multiple sets of random variables ${\bf X}_i \in {\bf R}^{I_i \times J}$ ($i=1,2,\ldots,N$) that are normalized to have zero mean and unit variance. The MAXVAR objective function for maximization of the overall correlation among canonical variates from them is defined as
\begin{eqnarray}\label{E-MsetCCA}
\mathop{\max}\limits_{{\bf w}_1,\ldots,{\bf w}_N} & \rho = \sum\limits_{i\neq j}^N {\bf w}_i^T{\bf X}_{i}{\bf X}_{j}^T{\bf w}_j    \nonumber   \\
s.t. & ~\frac{1}{N}\sum\limits_{i=1}^N{\bf w}_i^T{\bf X}_{i}{\bf X}_{i}^T{\bf w}_i=1,
\end{eqnarray}
With the method of Lagrange multipliers, the maximization in (\ref{E-MsetCCA}) can be transformed into the following generalized eigenvalue problem
\begin{equation}
({\bf R}-{\bf S}){\bf w}=\rho{\bf S}{\bf w},
\end{equation}
where
\begin{eqnarray}
{\bf R} & = & \left[ {\begin{array}{*{20}{c}}
                        {\bf X}_{1}{\bf X}_{1}^T & \ldots & {\bf X}_{1}{\bf X}_{N}^T \\
                        \vdots & \ddots & \vdots  \\
                        {\bf X}_{N}{\bf X}_{1}^T & \ldots & {\bf X}_{N}{\bf X}_{N}^T
                        \end{array}} \right],    \nonumber   \\
{\bf S} & = & \left[ {\begin{array}{*{20}{c}}
                        {\bf X}_{1}{\bf X}_{1}^T & \ldots & 0 \\
                        \vdots & \ddots & \vdots  \\
                        0 & \ldots & {\bf X}_{N}{\bf X}_{N}^T
                        \end{array}} \right],   \nonumber \\
{\bf w} & = & \left[ {\begin{array}{*{20}{c}}
                          {\bf w}_1  \\
                          \vdots  \\
                          {\bf w}_N
                          \end{array}} \right].   \nonumber
\end{eqnarray}
Then, linear transforms ${\bf w}_1,{\bf w}_2,\ldots,{\bf w}_N$ resulting in the largest overall canonical correlation among the multiple canonical variates ${\bf \tilde z}_i={\bf w}_i^T{\bf X}_i$ $(i=1,2,\ldots,N)$ are given by eigenvectors corresponding to the largest generalized eigenvalue.

Assume ${\bf X}_{1,m},{\bf X}_{2,m},\ldots,{\bf X}_{N,m} \in {\bf R}^{C \times P}$ ($C$ channels $\times$ $P$ time points) denote EEG data sets recorded from $N$ experimental training trials at the $m$-th stimulus frequency $f_m$. MsetCCA is implemented to find multiple linear transforms (i.e., spatial filters) ${\bf w}_{1,m},{\bf w}_{2,m},\ldots,{\bf w}_{N,m}$ that result in the maximization of overall correlation among the canonical variates ${\bf \tilde z}_{1,m},{\bf \tilde z}_{2,m},\ldots,{\bf \tilde z}_{N,m}$, with the joint spatial filtering ${\bf \tilde z}_{i,m} = {\bf w}_{i,m}^T{\bf X}_{i,m}$ ($i=1,2,\ldots,N$). These obtained canonical variates represent the common features shared among multiple sets of training data, which are assumed to reflect more accurately the real SSVEP characteristics than the artificial sine-cosine waves do. The optimized reference signal set at stimulus frequency $f_m$ is then constructed by the combination of canonical variates as
\begin{equation} \label{E-ORS}
{\bf Y}_{m}=\left[ {{\bf \tilde z}_{1,m}^T,{\bf \tilde z}_{2,m}^T,\ldots,{\bf \tilde z}_{N,m}^T} \right]^T.
\end{equation}
After the aforementioned calibration procedure of reference signals optimization, the maximal correlation coefficient $\rho_m$ between a new test data set ${\bf \hat X} \in {\bf R}^{C \times P}$ and each of the optimized reference signal sets ${\bf Y}_m \in {\bf R}^{N \times P}$ $(m=1,2,\ldots,M)$ is computed by CCA in (\ref{E-CCA}). The SSVEP frequency is then recognized according to (\ref{E-Ft}). Fig.~\ref{F-MsetCCAforSSVEP} illustrates the proposed frequency recognition method based on MsetCCA. Matlab code is available on request.

\vspace*{-1pt}
\subsection{MwayCCA for SSVEP Recognition}
\vspace*{-4pt}
One method of multiway canonical correlation analysis (MwayCCA) introduced in our previous study \cite{MCCA,L1MCCA} also considers to improve the recognition accuracy of SSVEP frequency through a calibration procedure of reference signal optimization. Different from the MsetCCA method proposed in this study, the MwayCCA method implements reference signal optimization by collaboratively maximizing correlation between the multiple dimensions of EEG tensor data and the pre-constructed sine-cosine waves. Consider a three-way tensor ${\cal X}_m =({\cal X})_{i_1,i_2,i_3} \in {\bf R}^{C \times P \times N}$ ($C$ channels $\times$ $P$ time points $\times$ $N$ trials) constructed by EEG data from $N$ experimental training trials at the $m$-th stimulus frequency ($m=1,2,\ldots,M$) and an original reference signal set ${\bf Y}_m \in {\bf R}^{2H \times P}$ constructed as (\ref{E-SinCos}). The MwayCCA method seeks three linear transforms ${\bf w}_{1,m} \in {\bf R}^C$, ${\bf w}_{3,m} \in {\bf R}^N$ and ${\bf v}_m \in {\bf R}^{2H}$ to maximize the correlation between linear combinations ${\bf \tilde z}_m = {\cal X}_m \times_1 {\bf w}_{1,m}^T \times_3 {\bf w}_{3,m}^T$ and ${\bf \tilde y}_m = {\bf v}_m^T{\bf Y}_m$ as
\begin{equation} \label{E-MCCA}
\mathop{\max}\limits_{{\bf w}_{1,m},{\bf w}_{3,m},{\bf v}_m} \rho_m = \frac{E\left[{{\bf \tilde z}_m{\bf \tilde y}_m^T}\right]}{\sqrt{E\left[{{\bf \tilde z}_m{\bf \tilde z}_m^T}\right]E\left[{{\bf \tilde y}_m{\bf \tilde y}_m^T}\right]}}.
\end{equation}
${\cal X} \times_k {\bf w}^T$ denotes the $k$-th way projection of a tensor with a vector and its definition can be found in literature \cite{MCCA}. The optimization problem in (\ref{E-MCCA}) can be solved by alternating CCAs \cite{MCCA}.

After obtaining the optimal linear transforms, we compute the canonical variate ${\bf \tilde z}_m$ as the optimized reference signal at stimulus frequency $f_m$. The maximal correlation coefficient $\rho_m$ between a new test data set ${\bf \hat X} \in {\bf R}^{C \times P}$ and each of the optimized reference signal ${\bf \tilde z}_m$ ($m=1,2,\ldots,M$) is computed by CCA in (\ref{E-CCA}). The SSVEP frequency of a test set is then recognized by (\ref{E-Ft}) again.

\begin{figure*}[!t]
\vspace*{13pt}
\centerline{\psfig{file=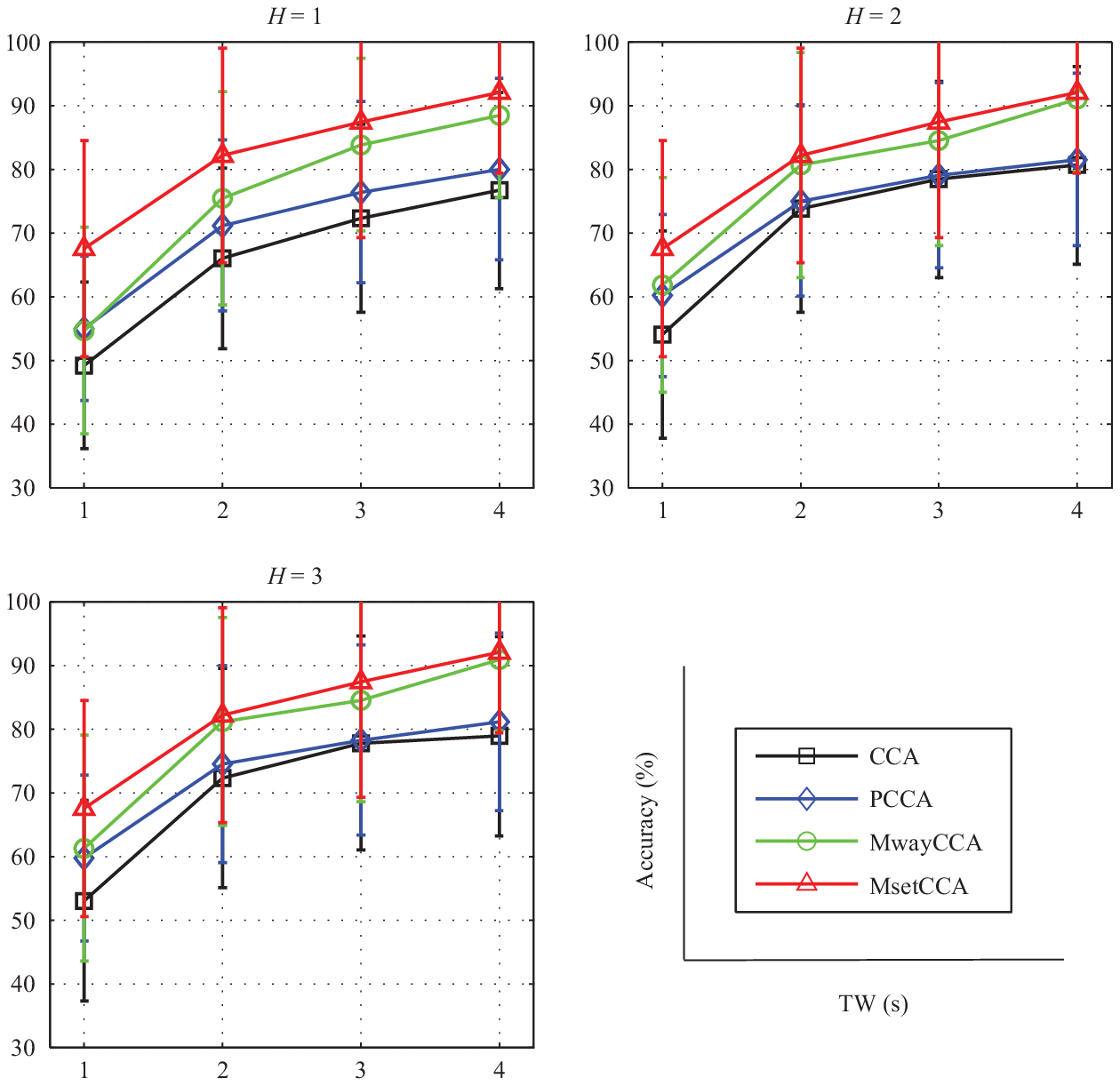,width=0.7\textwidth}}
\vspace*{13pt}
\fcaption{Averaged SSVEP recognition accuracies derived by the CCA, PCCA, MwayCCA and MsetCCA methods, respectively, with the number of harmonics $H=1,2,3$ and time window lengths (TWs) from 1 s to 4 s. Here eight channels P7, P3, Pz, P4, P8, O1, Oz and O2 were used. The errorbar denotes the standard deviation of accuracy on all the ten subjects. Note that recognition accuracies obtained with $H=1,2,3$ are exactly the same as each other for the MsetCCA method, since it does not require pre-defined $H$ but automatically estimate the optimal SSVEP features from the training data.} \label{F-HarmonicAccuracy}
\end{figure*}
\vspace*{13pt}

\vspace*{-1pt}
\subsection{PCCA for SSVEP Recognition}
\vspace*{-4pt}
On the other hand, a phase constrained CCA method (PCCA) \cite{SSVEP-PCCA} has been recently proposed for SSVEP frequency recognition. In the PCCA method, SSVEP response phase that estimated from the apparent latency of training data is adopted as a constraint on the pre-constructed reference signals in the CCA method. The constrained reference signals more effectively captures the phase information of SSVEP frequency, and hence significantly improves recognition accuracy of the CCA method.

The SSVEP response phase $\phi_r$ is proportional to the stimulus frequency
\begin{equation}\label{E-Phir}
\phi_r(f_m) = -L \cdot f_m \cdot 360^{\circ},
\end{equation}
where $L$ denotes the apparent latency that is fixed for all the stimulus frequencies $f_m$ ($m=1,2,\ldots,M$) and can be estimated from the EEG training data based on SSVEP phase $\phi_s$ \cite{PhaseLatency}. The SSVEP phase $\phi_s(f_m)$ at the stimulus frequency $f_m$ is computed by discrete Fourier transform of EEG signal from channel Oz at $f_m$. The difference between $\phi_s$ and $\phi_r$ is a multiple of $360^{\circ}$ as
\begin{equation} \label{E-PhirPhis}
\phi_r(f_m) = \phi_s(f_m)-n(f_m) \cdot 360^{\circ},
\end{equation}
where $n(f_m)$ is an integer depending on the stimulus frequency $f_m$. According to (\ref{E-Phir}) and (\ref{E-PhirPhis}), we obtain $n(f_m) = L \cdot f_m + \phi_s(f_m)/360^{\circ}$. The apparent latency $L$ is then determined through an exhaustive search procedure based on weighting least-squares fit. With the estimated $L$, the SSVEP response phase $\phi_r(f_m)$ $(m=1,2,\ldots,M)$ can be derived from (\ref{E-Phir}). The Optimized reference signal set at stimulus frequency $f_m$ is obtained by adding $\phi_r(f_m)$ as a constraint to the pre-constructed sine-cosine waves. The SSVEP frequency of a test set is then recognized by using (\ref{E-CCA}) and (\ref{E-Ft}) again. More details about the PCCA method can be found in literature \cite{SSVEP-PCCA}.

\begin{table*}[htbp]
\tcap{Statistical analysis results of accuracy difference between the MsetCCA and each of the CCA, PCCA, MwayCCA with different numbers of harmonics $H$ at various time window lengths (TWs), respectively.}
\vglue-6pt
\centerline{\small\baselineskip=13pt
\begin{tabular}{l l l l l}\\
\hline
Method Comparison & TW &    &    &      \\
                                  &  1 s  &  2 s  &  3 s  &  4 s   \\
\hline
$H=1$    &   &    &    &    \\
MsetCCA vs. CCA    &  $p<0.001$  &  $p<0.001$  &  $p<0.001$  &  $p<0.001$  \\
MsetCCA vs. PCCA    &  $p<0.01$  &  $p<0.05$  &  $p<0.01$  &  $p<0.005$  \\
MsetCCA vs. MwayCCA    &  $p<0.05$  &  $p<0.05$  &  $p=0.28$  &  $p<0.05$  \\
$H=2$    &   &    &    &    \\
MsetCCA vs. CCA    &  $p<0.005$  &  $p<0.05$  &  $p<0.05$  &  $p<0.005$  \\
MsetCCA vs. PCCA    &  $p<0.05$  &  $p<0.01$  &  $p<0.05$  &  $p<0.005$  \\
MsetCCA vs. MwayCCA    &  $p=0.19$  &  $p=0.46$  &  $p=0.21$  &  $p=0.51$  \\
$H=3$    &   &    &    &    \\
MsetCCA vs. CCA    &  $p<0.005$  &  $p<0.005$  &  $p<0.005$  &  $p<0.001$  \\
MsetCCA vs. PCCA    &  $p<0.01$  &  $p<0.005$  &  $p<0.05$  &  $p<0.005$  \\
MsetCCA vs. MwayCCA    &  $p=0.15$  &  $p=0.61$  &  $p=0.23$  &  $p=0.46$  \\
\hline\\
\end{tabular}}\label{Tab-HAcc}
\end{table*}

\vspace*{-1pt}
\subsection{Experimental Evaluation}
\vspace*{-4pt}
In this study, the proposed MsetCCA method is compared with the CCA method \cite{SSVEP-CCA}, the MwayCCA method \cite{MCCA} and the PCCA method \cite{SSVEP-PCCA}, to validate its effectiveness for SSVEP frequency recognition.

Since the occipital and parietal scalp areas have been demonstrated to contribute most to the SSVEP frequency recognition \cite{SSVEP-CCA,SSVEP-Bin}, only eight channels P7, P3, Pz, P4, P8, O1, Oz and O2 are used for analysis in this study.

For the MsetCCA, the MwayCCA and the PCCA methods, leave-one-run-out cross-validation is implemented to evaluate the average recognition accuracy. More specifically, the data from 19 runs (4 trials in each run) are used as training data for reference signal optimization while the data from the left-out run for validation. This procedure is repeated for 20 times such that each run serves once for validation. For the CCA method, the average recognition accuracy is evaluated on the direct validation of 20 runs.

\begin{figure*}[!t]
\vspace*{13pt}
\centerline{\psfig{file=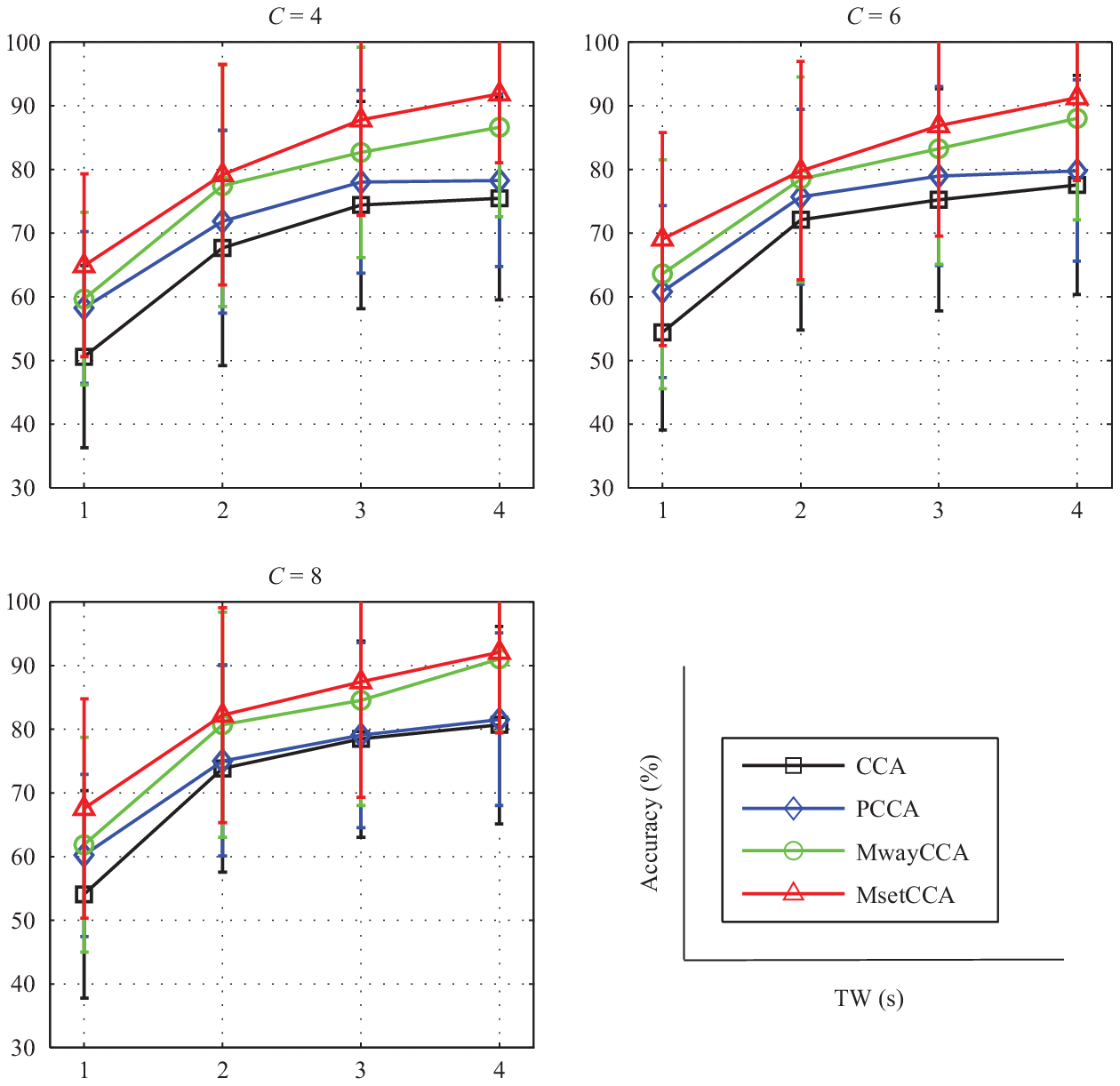,width=0.7\textwidth}}
\vspace*{13pt}
\fcaption{Averaged SSVEP recognition accuracies derived by the CCA, PCCA, MwayCCA and MsetCCA methods, respectively, with the number of channels $C=4,6,8$ and time window lengths (TWs) from 1 s to 4 s. $C=4$: P3, P4, O1 and O2; $C=6$: P7, P3, P4, P8, O1 and O2; $C=8$: P7, P3, Pz, P4, P8, O1, Oz and O2. Here the number of harmonics $H=2$. The errorbar denotes the standard deviation of accuracy on all the ten subjects.} \label{F-ChannelAccuracy}
\end{figure*}
\vspace*{13pt}

\section{Results}
\noindent
Since the number of harmonics $H$ is required to be pre-defined for the CCA, the PCCA and the MwayCCA methods, we first investigate effects of varying $H$ on the frequency recognition accuracy. Fig.~\ref{F-HarmonicAccuracy} shows the averaged SSVEP recognition accuracies obtained by the CCA, the PCCA, the MwayCCA and the MsetCCA methods with $H$ from 1 to 3 at various time window lengths (TWs), respectively. Note that the proposed MsetCCA method did not require the pre-defined $H$, and hence derived the exactly same accuracy for $H=1,2$ and 3. The statistical analysis results of accuracy differences evaluated by paired-sample t-test are summarized in Table~\ref{Tab-HAcc}. For $H=1,2$ and 3, the MsetCCA method achieved the best frequency recognition accuracy at all the four TWs. The CCA, the PCCA and the MwayCCA methods yielded higher accuracies with $H=2$ and 3 in contrast to $H=1$ while no big accuracy changing was found for the three methods in increasing $H$ from 2 to 3. Thus, we choose $H=2$ for the CCA, the PCCA and the MwayCCA methods in further analysis.

With fixed $H=2$, effects of varying the number of channels $C$ on the frequency recognition accuracy are further studied. Fig.~\ref{F-ChannelAccuracy} shows averaged SSVEP recognition accuracies obtained by the four methods with $C=4,6$ and 8 at various time window lengths (TWs), respectively. Table~\ref{Tab-CAcc} summarizes the statistical analysis results of accuracy differences evaluated by paired-sample t-test. For $C=4,6$ and 8, the MsetCCA method derived better frequency recognition accuracies than the other three methods, especially when using a small number of channels and a short TW. For instance, the MsetCCA method significantly outperformed the CCA, the PCCA and the MwayCCA methods with $C=4$ at the TW of 1 s (MsetCCA $>$ CCA: $p<0.001$, MsetCCA $>$ PCCA: $p<0.005$, MsetCCA $>$ MwayCCA: $p<0.05$).

\begin{figure*}[!t]
\centerline{\psfig{file=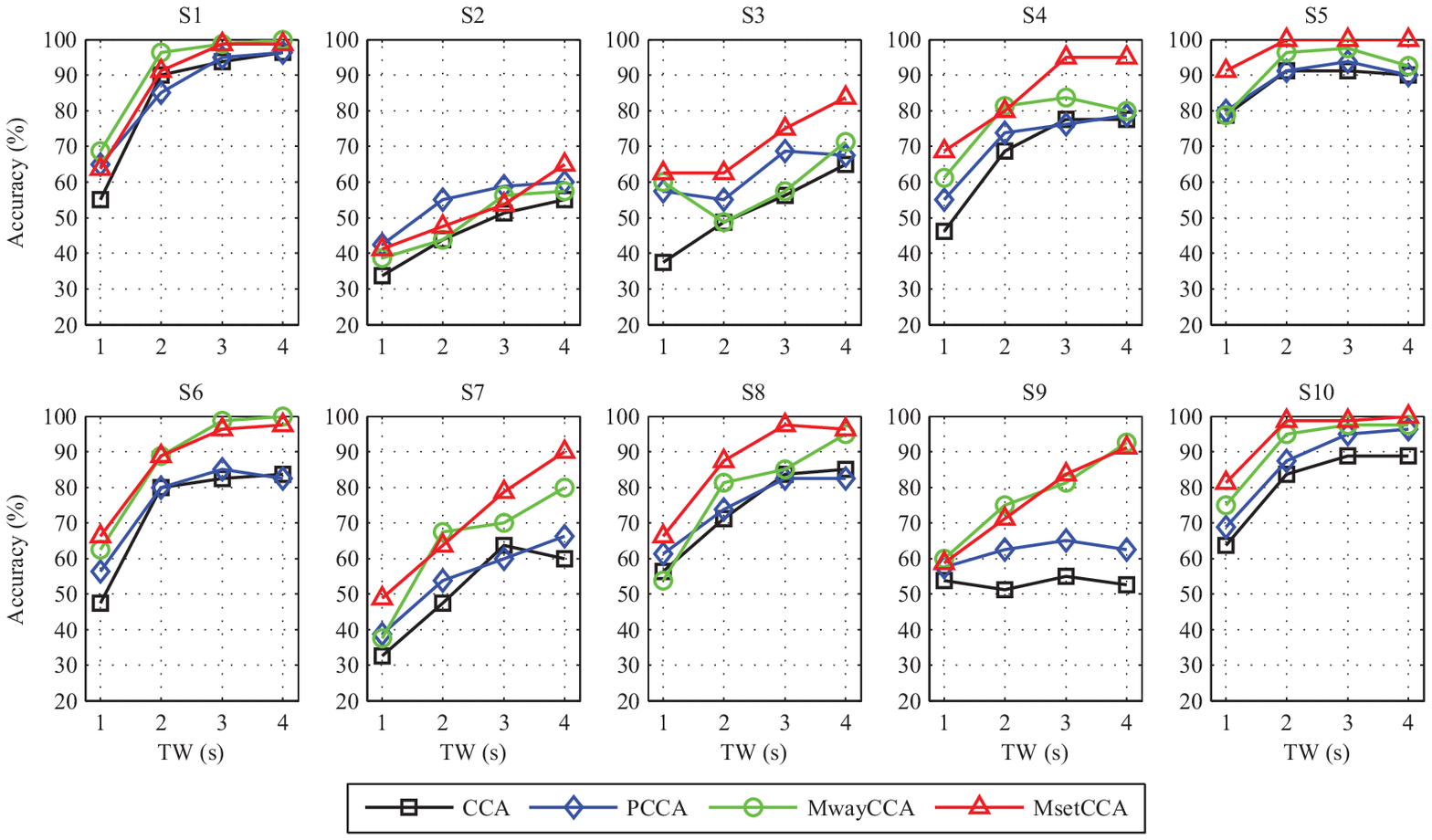,width=0.7\textwidth}}
\vspace*{13pt}
\fcaption{SSVEP recognition accuracies derived by the CCA, PCCA, MwayCCA and MsetCCA methods, respectively, for each of the ten subjects. Here the number of harmonics $H$ is 2 for the CCA, the PCCA, the MwayCCA methods, the number of channels $C$ is 4. The averaged accuracies is shown in the subfigure with $C=4$ of Fig.~\ref{F-ChannelAccuracy}.} \label{F-SubjectwiseAccuracy}
\end{figure*}

\begin{figure*}[!t]
\centerline{\psfig{file=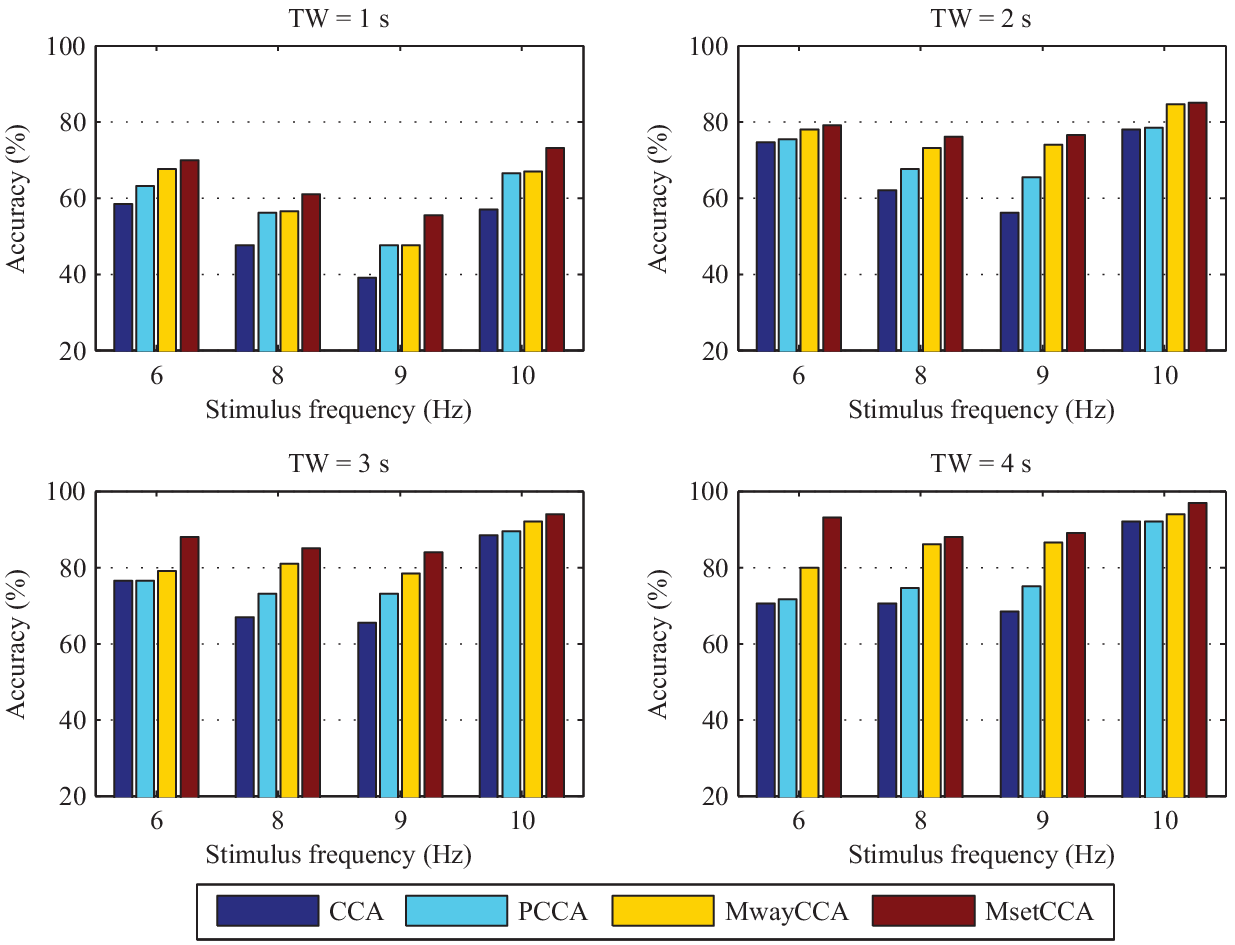,width=0.7\textwidth}}
\vspace*{13pt}
\fcaption{SSVEP recognition accuracies obtained by the CCA, PCCA, MwayCCA and MsetCCA methods, respectively, with different time window lengths (TWs), for each of the four stimulus frequencies, averaged on all subjects. Here the number of harmonics $H$ is 2 for the CCA, the PCCA, the MwayCCA methods, the number of channels $C$ is 4.} \label{F-PerFrequencyAccuracy}
\end{figure*}

Fig.~\ref{F-SubjectwiseAccuracy} depicts the subject-wise accuracy with the number of channels $C=4$. For most subjects, the MsetCCA method yielded higher accuracy than the other three methods at various TWs. Fig.~\ref{F-PerFrequencyAccuracy} further shows the SSVEP recognition accuracy averaged on all subjects at each of the four stimulus frequencies for the four methods, respectively. For all of the four stimulus frequencies, the MsetCCA method consistently outperformed the other three competing methods.

In summary, the aforementioned results indicate that the proposed MsetCCA method is promising for the development of SSVEP-based BCIs with high performance.

\begin{table*}[htbp]
\tcap{Statistical analysis results of accuracy difference between the MsetCCA and each of the CCA, PCCA, MwayCCA with different numbers of channels $C$ at various time window lengths (TWs), respectively.}
\vglue-6pt
\centerline{\small\baselineskip=13pt
\begin{tabular}{l l l l l}\\
\hline
Method Comparison & TW &    &    &      \\
                                  &  1 s  &  2 s  &  3 s  &  4 s   \\
\hline
$C=4$    &   &    &    &    \\
MsetCCA vs. CCA    &  $p<0.001$  &  $p<0.005$  &  $p<0.005$  &  $p<0.005$  \\
MsetCCA vs. PCCA    &  $p<0.005$  &  $p<0.005$  &  $p<0.005$  &  $p<0.005$  \\
MsetCCA vs. MwayCCA    &  $p<0.05$  &  $p=0.36$  &  $p<0.05$  &  $p<0.05$  \\
$C=6$    &   &    &    &    \\
MsetCCA vs. CCA    &  $p<0.001$  &  $p<0.01$  &  $p<0.005$  &  $p<0.01$  \\
MsetCCA vs. PCCA    &  $p<0.01$  &  $p=0.086$  &  $p<0.05$  &  $p<0.005$  \\
MsetCCA vs. MwayCCA    &  $p<0.05$  &  $p=0.67$  &  $p=0.10$  &  $p<0.05$  \\
$C=8$    &   &    &    &    \\
MsetCCA vs. CCA    &  $p<0.001$  &  $p<0.05$  &  $p<0.05$  &  $p<0.005$  \\
MsetCCA vs. PCCA    &  $p<0.005$  &  $p<0.01$  &  $p<0.05$  &  $p<0.005$  \\
MsetCCA vs. MwayCCA    &  $p<0.05$  &  $p=0.46$  &  $p=0.21$  &  $p=0.51$  \\
\hline\\
\end{tabular}}\label{Tab-CAcc}
\end{table*}

\vspace*{13pt}

\vspace*{13pt}

\vspace*{-25pt}
\section{Discussion}
\begin{figure*}[!t]
\vspace*{13pt}
\centerline{\psfig{file=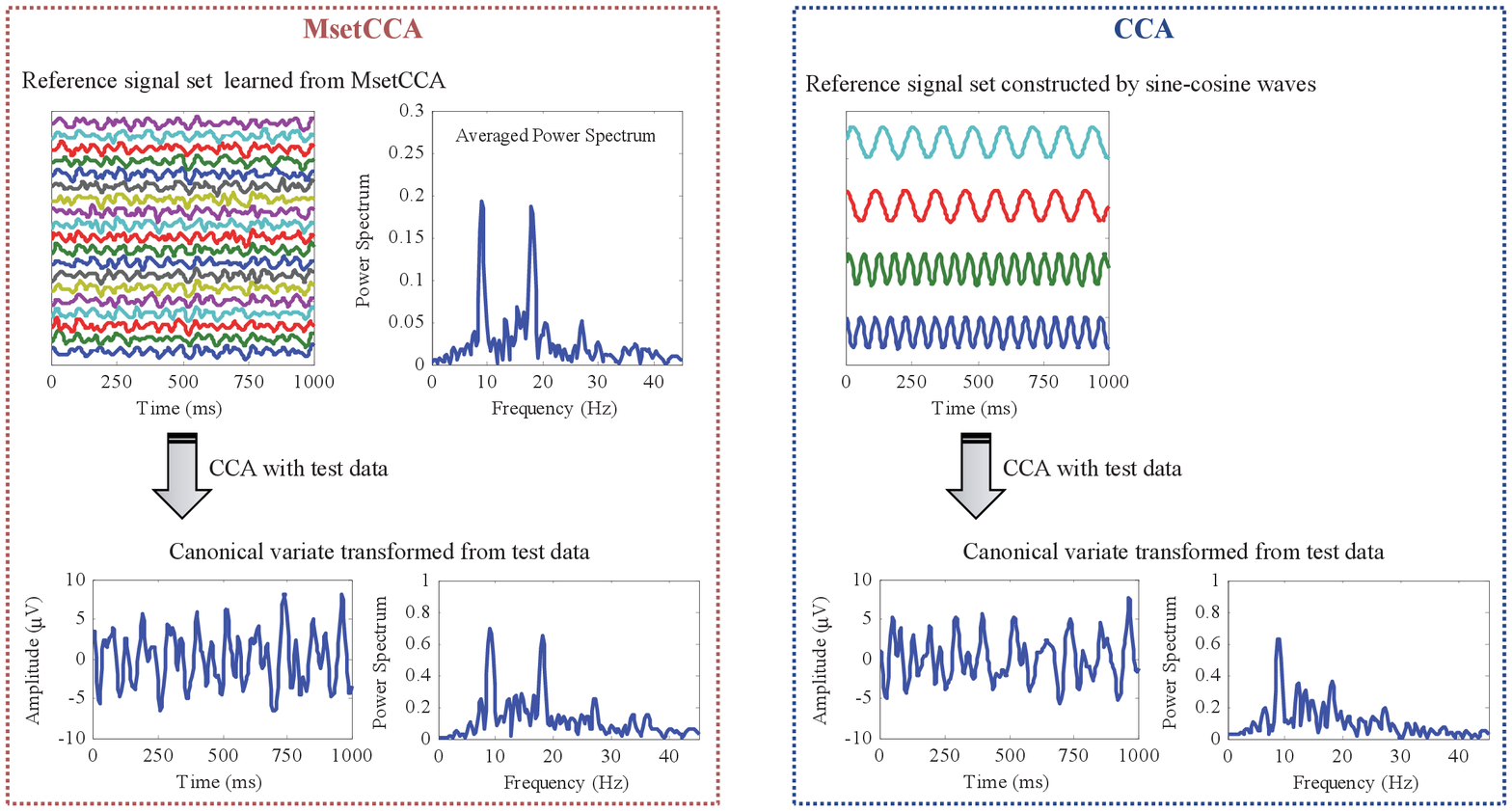,width=0.85\textwidth}}
\vspace*{13pt}
\fcaption{Illustration for superiority of the reference signals learned from MsetCCA over the sine-cosine waves for SSVEP recognition. The reference signals (i.e., common features) extracted by MsetCCA more accurately captured the harmonic feature in test data than the sine-cosine waves did.} \label{F-PloRef}
\end{figure*}

\noindent
For EEG classification or detection, many researches have confirmed that a sophisticated calibration with appropriate analysis method could significantly improve the accuracy \cite{PFD}$^-$\cite{CongERP3}. In this study, instead of directly using pre-constructed sine-cosine waves as reference signals in the CCA method for SSVEP frequency recognition, all the PCCA, the MwayCCA and the MsetCCA methods consider to improve the accuracy through optimizing the reference signals from training data. The optimization procedure of reference signals in the MsetCCA method is completely based on the training data, whereas those in both the PCCA and the MwayCCA methods still need to resort to the pre-constructed sine-cosine waves. The MsetCCA method learns multiple linear transforms through the maximization of overall correlation among canonical variates, which provides a novel approach to extract SSVEP common features from the joint spatial filtering of multiple experimental trials at the same stimulus frequency. Recently, the positive impact of common features on classification has been suggested by Zhou et al. \cite{CIFA-Zhou}. In our study, the extracted common features are combined to form the optimized reference signals for robust recognition of SSVEP frequency. As an example, Fig.~\ref{F-PloRef} illustrates superiority of the reference signals learned from MsetCCA over the sine-cosine waves for SSVEP recognition. The reference signal set constructed by the common features more accurately captured the harmonic feature in test data than the sine-cosine waves did. This characteristic assisted in accurately recognizing the SSVEP frequency.

\vspace*{13pt}

\vspace*{-10pt}
Another important advantage of the MsetCCA method is that it does not require the pre-defined number of harmonics $H$. For the CCA, the PCCA and the MwayCCA methods, the pre-defined $H$ is necessary since their recognition procedures are based on the pre-constructed sine-cosine waves. Bin et al. \cite{SSVEP-Bin} reported that the second and third harmonics of SSVEP had no significant impact on the recognition accuracy, whereas M\"{u}ller-Putz et al. \cite{SSVEP-Muller} confirmed that the use of higher harmonics as features indeed improved the recognition accuracy. Some existing SSVEP-based BCIs \cite{MultiChannel-Friman,SSVEP-CCA,SSVEP-Zhang} adopted more than one harmonic (i.e., fundamental frequency) for SSVEP frequency recognition while some others used the fundamental frequency only \cite{PcodeSSVEP,SSVEP-SC,SSVEP-PCCA}. It can be seen that decision of the optimal $H$ is still based on experience, which may hardly give the best recognition accuracy for all subjects. In this study, the proposed MsetCCA method provided an effective optimization procedure for automatical estimation of the subject-specific SSVEP features with accurate harmonics from multi-trial EEG training data.

It is worth noting that reference signal optimization using the MsetCCA method can be efficiently implemented without complicated pre-processing. Under the computation environment of Matlab R2011a on a laptop with 1.20 GHz CPU, the computational time of optimization procedure is 0.357 s for the MsetCCA method, 1.475 s for the MwayCCA method and 3.091 s for the PCCA methods method, when using a TW of 4 s on four candidate stimulus frequencies. The computational cost is negligible in contrast to the time cost for training data recording. After optimizing the reference signals, SSVEP frequency recognition based on CCA can be efficiently executed in 0.0056 s.

The number of reference signals learned from the proposed MsetCCA method depends on the number of trials used in the training stage. Use of insufficient training trials could hardly learn effective reference signals, whereas use of too many training trials could introduce unavoidable redundancy into reference signals as well as additional memory and computational requirements for SSVEP recognition. To investigate this potential problem, Fig.~\ref{F-AccuracyDifTrainTim} shows the averaged accuracy and computational time for SSVEP recognition in using CCA with different numbers of training trials for reference signal optimization by MsetCCA. The overall trend of accuracy was improved with the increase in the number of training trials. However, the accuracy did not change too much when more than ten training trials were used. This result implies an appropriate selection of the number of reference signals (i.e., the number of training trials) could effectively reduce the requirement for training data without significant decrease in recognition accuracy. Hence, how to decide the optimal number of reference signals in the MsetCCA method is worthy of our further study. On the other hand, the computational time was still in several milliseconds due to the high efficiency of CCA although it increased with the increase in the number of training trials. Therefore, the computational time of recognition procedure in the MsetCCA method is negligible with an appropriate number of training trials.

In summary, with a sophisticated calibration of reference signals from training data, the proposed MsetCCA method significantly improved the recognition accuracy of SSVEP frequency in contrast to the popularly used CCA method. It should be noted that the CCA method is still preferred for development of a zero-training SSVEP-based BCI when training time reduction is considered more important than recognition accuracy improvement, since it does not require training data for reference signal optimization. However, we strongly prefer the MsetCCA method in development of an improved SSVEP-based BCI with high performance, especially when the recognition accuracy is considered as the primary factor.

\vspace*{13pt}
\centerline{\psfig{file=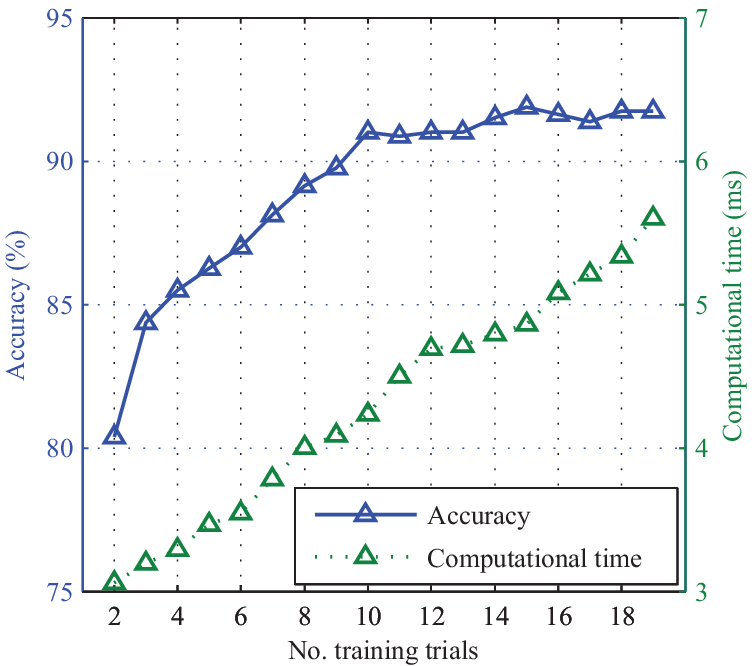,width=0.42\textwidth}}
\fcaption{Averaged accuracy and computational time for \\
SSVEP recognition in using CCA with different numbers \\
of reference signals learned from MsetCCA (i.e., by using \\
different numbers of training trials). Here, the number \\
of channels $C=4$, time window length (TW) is 4 s.} \label{F-AccuracyDifTrainTim}
\vspace*{13pt}

\section{Conclusions}
\noindent
In this study, we introduced a novel method based on multiset canonical correlation analysis (MsetCCA) to improve the SSVEP frequency recognition for BCI application. The MsetCCA method learned multiple linear transforms through maximizing the overall correlation among canonical variates to extract common features from the joint spatial filtering of multiple sets of EEG data recorded at the same stimulus frequency. Such common features reflected more accurately the SSVEP characteristics and were used instead of pre-constructed sine-cosine waves to form the optimized reference signals in the CCA method for SSVEP frequency recognition. Experimental results with EEG data from ten healthy subjects demonstrated that the proposed MsetCCA method outperformed the CCA method at the cost of using training data, and also performed better than the other two competing methods, i.e., MwayCCA and PCCA. Future study will investigate effectiveness of the MsetCCA method on a wide range of stimulus frequencies.

\nonumsection{Acknowledgements}
\noindent
The authors sincerely thank the editor and the anonymous reviewers for their insightful comments and suggestions that helped improve the paper. This study was supported in part by the Nation Nature Science Foundation of China under Grant 61305028, Grant 61074113, Grant 61203127, Grant 61103122, Grant 61202155, Fundamental Research Funds for the Central Universities Grant WH1314023, Grant WH1114038, and Shanghai Leading Academic Discipline Project B504.

\nonumsection{References}
\vspace*{-12pt}	

\end{multicols}
\end{document}